\documentclass[runningheads]{llncs}
\usepackage{amsfonts}
\usepackage{mathrsfs}
\usepackage{multirow}
\usepackage{color}
\usepackage{epsfig}
\usepackage{subfigure}
\usepackage{hyperref}
\usepackage{graphicx}
\usepackage{amsmath,amssymb} 
\usepackage{color}

\begin{document}
\title{Interaction-aware Spatio-temporal Pyramid Attention Networks for Action Classification} 

\titlerunning{ISTPAN for Action Classification}
%
\author{Yang Du\inst{1,2,3}\and
Chunfeng Yuan\inst{2}\thanks{Corresponding Author} \and
Bing Li\inst{2}\and 
Lili Zhao\inst{3} \and \\
Yangxi Li\inst{4} \and 
Weiming Hu\inst{2}}
%
\authorrunning{Yang Du et al.}
%

\institute{University of Chinese Academy of Sciences \and
CAS Center for Excellence in Brain Science and Intelligence Technology, National Laboratory of Pattern Recognition, Institute of Automation, CAS \and
Meitu, $^4~~$National Computer network Emergency Response technical
Team/Coordination Center of China\\
\email{duyang2014@ia.ac.cn,\{cfyuan,bli,wmhu\}@nlpr.ia.ac.cn,\\lili.zhao@meitu.com, liyangxi@outlook.com}}
\maketitle              
\begin{abstract}
Local features at neighboring spatial positions in feature maps have high correlation since their receptive fields are often overlapped. Self-attention usually uses  the weighted sum (or other functions) with internal elements of each local feature to obtain its weight score, which ignores interactions among local features. To address this, we propose an effective interaction-aware self-attention model inspired by PCA to learn attention maps. Furthermore, since different layers in a deep network capture feature maps of different scales, we use these feature maps to construct a spatial pyramid and then utilize multi-scale information to obtain more accurate attention scores, which are used to weight the local features in all spatial positions of feature maps to calculate attention maps. Moreover, our spatial pyramid attention is unrestricted to the number of its input feature maps so it is easily extended to a spatio-temporal version. Finally, our model is embedded in general CNNs to form end-to-end attention networks for action classification. Experimental results show that our method achieves the state-of-the-art results on the UCF101, HMDB51 and untrimmed Charades.
\end{abstract}
\section{Introduction}
Human action recognition \cite{ref51,ref52,ref53,ref54,ref37,ref55} in videos occupies a significant position in computer vision and has attracted a large amount of attention. The CNN-based methods \cite{ref43,ref44,ref45,ref46} achieve great progress in image classification. Besides, there are more labeled images to train the networks than labeled video data. In view of these two points, many methods combine predictions of images from a video by image-based classification methods to classify videos. However, videos own not only much irrelevant information with human actions in intra frames but also include more temporal information along frames.

Attention \cite{ref78,ref75} enables  models to differentiate irrelevant information so as to focus on the key information. Hard and soft attention are two typical ways. Hard attention usually needs to make hard binary choices and faces training problems while soft attention uses weighted average instead of hard selection. In action recognition, many existing methods \cite{ref74,ref77} often use CNN or RNN based architectures to extract the channel-level or frame-level local features, which are subsequently modeled by combining LSTM with soft attention. Usually, supplementary sources are required to predict the attention scores of the next time by LSTM, so substantial computational cost is required.  Self-attention \cite{ref75} is one kind of soft attention and can reduce computational cost by performing on top of features without extra input, and it is also a special form of the non-local network \cite{ref79} by using attention scores to weight all features to obtain salient features. Local features at neighboring spatial positions in feature maps have high correlation since their receptive fields are often overlapped. However, the interaction information among features is often ignored in self-attention because the attention score of each feature is usually calculated by the weighted sum (or other functions) of internal elements of this feature.

We propose an interaction-aware spatio-temporal pyramid attention layer. It is embedded in general CNNs to generate more discriminative attention networks for video action classification. Firstly, attention enables models to extract key features with high attention scores, just as that PCA extracts key features with principal components. By minimizing the trace of the covariance matrix, PCA utilizes the interaction information among features to obtain basis vectors for projection. Inspired by PCA, we propose an interaction-aware self-attention by using interaction information to train their attention scores. Secondly, considering that feature pyramid \cite{ref80,ref81} provides an important advantage of the multi-resolutions for feature representation, we stack feature maps of different layers to construct a spatial pyramid. Then we perform an interaction-aware self-attention on the channel-level features in the pyramid to obtain more accurate attention scores, which are used to aggregate the top layer of the pyramid to obtain more discriminative attentional maps. Thirdly, spatio-temporal detection \cite{ref7,ref12} based methods also have good performance so it is promising to use attention to detect salient spatio-temporal regions in videos. So we extend the spatial pyramid attention to a spatio-temporal version to detect and utilize the key information in videos. Besides, our model is irrelevant to the temporal sequence order and is compatible to frames of any number.

\textbf{Contributions: } (1) We propose an interaction-aware self-attention inspired by PCA. (2) We propose to construct a spatial feature pyramid to obtain more accurate attention maps by multi-scale information. (3) We extend our layer to a spatio-temporal version which accepts frames of any number even though the architecture and parameters of our layer are determined. (4) Our layer is able to apply in any CNNs to form end-to-end attention networks. Finally, we validate our attention networks generated from three baseline networks, VGGNet-16, BN-Inception, and Inception-ResNet-V2, on the UCF101, HMDB51 and untrimmed Charades datasets, and obtain the state-of-the-art results. 
\section{Related Works}
\textbf{Deep Networks:} CNN-based methods have obtained great progress in image recognition compared with hand-crafted features \cite{ref73,ref8,ref9,ref23,ref11,ref22}. Two-Stream ConvNet \cite{ref20} first used the image-based recognition method to solve the issue of video classification by averaging classification scores of uniformly sampled frames from a video as the video prediction. Deep ConvNets \cite{ref28} used the fusion of different layers and were trained on a large scale dataset such as Sports-1M. Temporal Segment Network \cite{ref48} divided the video into multiple parts, on which Two-Stream ConvNets were used separately.  ST-ResNet \cite{ref66} based on ResNet \cite{ref43} also employed two-stream architecture. The 3D ConvNet \cite{ref19,ref93,ref94} extended the 2D ConvNet to directly train using videos, but it needed abundant computations and more pre-training on larger dataset such as Kinetics \cite{ref94}. The works \cite{ref42,ref61} explored video representations based on spatio-temporal convolutions. These methods equally treated the information of each frame or spatio-temporal region. The performance is limited because it is hard to effectively differentiate key features. Many deep networks are used to model the temporal structure, such as the recurrent neural network (RNN) \cite{ref29} or its variants such as long short-term memory (LSTM) \cite{ref30,ref31,ref32,ref62} or CNNs \cite{ref62,ref68}.  

\textbf{Attention Methods:} Hard attention usually needs to add extra information to enhance the original model. \cite{ref82,ref83} proposed attention RNNs for objection recognition to select regions by making hard binary choices. R*CNN \cite{ref84} used an auxiliary box to encode context besides the human bounding box. \cite{ref86}  used the whole image as the context and used multiple instance learning (MIL) to note all humans in the image to predict the action label for the input image. Soft attention uses weight average instead of hard selection. Sharma $et~al.$ \cite{ref74} proposed a soft-attention LSTM on top of the RNNs to pay attention to salient parts of the video frames for classification. Li $et~al.$ \cite{ref85} proposed an end-to-end sequence learning model called VideoLSTM. However, these soft attention models also required the auxiliary information to guide the weight average. Girdhar and Ramanan \cite{ref87} proposed the unconstrained self-attention pooling added at the last layer to generate a global representation on top of a CNN. Long $et~al.$ \cite{ref76} proposed a method of attention clusters to integrate local feature of each frame by self-attention.  Ma $et~al.$ \cite{ref88} proposed a model by attending to key image-level representations to summarize the whole video sequence with LSTM. Previous attention based methods often focused on the frame-level deep network and the average performance was limited. We propose an interaction-aware self-attention method to weight the channels of feature maps, and use the feature maps of different scales to construct spatial pyramid to obtain more accurate attention scores. Finally, the temporal extension of our layer can be adaptive to the video-level framework for action classification.
\section{Interaction-aware Spatio-temporal Pyramid Attention}
\label{sec:spatio-temporal Pyramid Attention}
In this section, we first propose an interaction-aware spatial pyramid attention layer inspired by PCA. This layer can be plugged into a general CNN to form an end-to-end attention network, in which this layer can generate more discriminative attention feature maps. Next, we extend our layer to a temporal version for aggregating temporal sequences for action classification in videos. Then, we give the modified loss function of our network. Finally, we give implemental details.
\subsection{Interaction-aware Spatial Pyramid Attention Layer}
\label{sec:sub:Spatial Pyramid Attention Layer}
The CNNs usually extract feature maps by equally treating every local region. We aim to adding an attention layer into the general CNN after one convolutional layer to emphasize the features of the key local regions and further improve the performance of the network. Let $f_i\in R^{W_{i} \times H_{i} \times C_{i}}$ denote a group of feature maps at the $i$-th layer of a network obtained by inputting a frame, where $W_{i}, H_{i}, C_{i}$ are the spatial size and channel number of the feature maps. We flatten $f_i$ into $X_i \in R^{W_iH_i \times C_i}$. $X_i$ can be considered as stacked rows $X_i=[{h^i}^T_1,{h^i}^T_2,...,{h^i}^T_{W_iH_i}]^T$, where $h_k^i\in R^{1\times C_i}$ is a channel vector locating at the $k$-th spatial position of $f_i$ and represents the local feature of its receptive field in the input image. We propose an interaction-aware spatial pyramid attention layer to generate discriminative $d$ features $M_i \in R^{d \times C_{i}}$ from local features $\{h^i_k\}_{k=1}^{W_iH_i}$. To preserve the architecture of the network behind the $i$-th layer unchanged, we set $d=W_iH_i$. Then, $M_i$ is reshaped into  attention maps $M_i^{'}\in R^{W_{i} \times H_{i} \times C_{i}}$.

\begin{figure}[t]
\setlength{\abovecaptionskip}{-0mm} 
\setlength{\belowcaptionskip}{-0mm} 
\centering
\includegraphics[width=118mm,height=45mm]{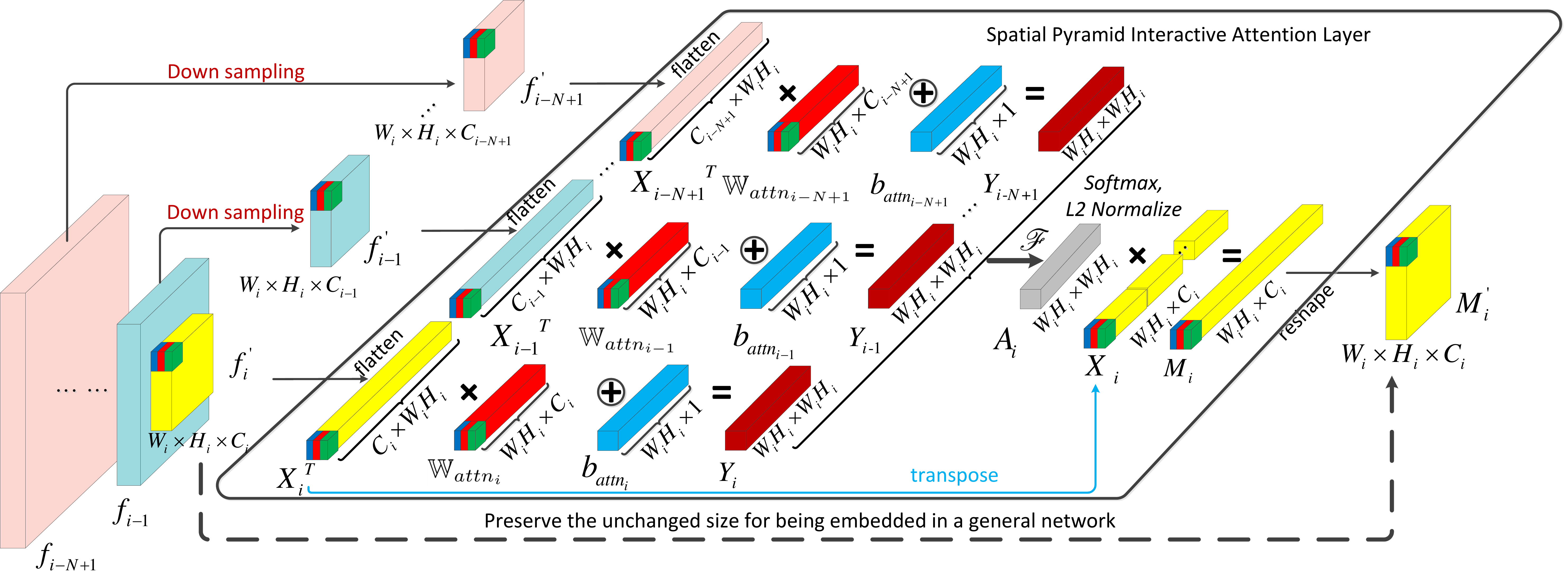}
\caption{The illustration of our spatial pyramid interactive attention layer. We use the feature maps of different sizes in different layers to construct a multi-scale attention to obtain more accurate spatial attention.}
\label{fig:Spatial Pyramid Attention Layer}
\end{figure}

Specifically, we use feature maps of the $i$-th layer and that before the $i$-th to construct a feature pyramid $\{f_j\in R^{W_j\times H_j\times C_j}\}|_{j=i-N+1}^{i}$, where $N$ is the number of pyramid layers, as shown in Figure \ref{fig:Spatial Pyramid Attention Layer}. Then, we down sample feature maps except the top layer of the pyramid to fit the spatial size of the top layer,
\begin{gather}
  f_j^{'}=\left\{ \begin{aligned}
         \Re(f_j),~~~j&=i-N+1,...,i-1,\\
         f_i,~~~~~~~~j&=i.~~~~~~~~~~~~~~~~~~~~~~
        \end{aligned} \right. ,where~f_j^{'}\in R^{W_i\times H_i\times C_j},
\label{equ:f_j}
\end{gather}
where $\Re(\bullet)$ is a down sampling function. To adapt to multiple feature maps $\{f_j^{'}\}|_{j=i-N+1}^i$ with different channels, we perform a self-attention on the channel vectors $h^j$. Namely, given $X_j$ flattened from $f_j^{'}$, we obtain an attention score matrix $Y_j$ by individually compute the sum of weighted channels of every $h^j$ in $X_j$, which is formulated as
\begin{gather}
  Y_j = \mathbb{W}_{attn_j}X_j^T\oplus b_{attn_j}, where~\mathbb{W}_{attn_j}\in R^{d\times C_j},~X^T_j\in R^{C_j\times W_iH_i},
\label{equ:X_j}
\end{gather}
where $\mathbb{W}_{attn_j}=\{w^j_m\in R^{1\times C_j}\}_{m=1}^{d}$  and $b_{attn_j}\in R^{d\times 1} =\{b^j_k\}_{k=1}^d$ are trainable weights and biases. The symbol $\oplus$ denotes that each column of $\mathbb{W}_{attn_j}X_j^T$ adds $b_{attn_j}$. Furthermore, let $Y_j=[{y^j_1}^T,{y^j_2}^T,...,{{y^j}^T_{d}}]^T$, where $y^j\in R^{1\times W_iH_i}$ denotes attention scores corresponding to all spatial positions of each $j$th-layer feature maps. Each score in $y^j$ is individually calculated by
$w^j_m{{h^j}^T_k}+b^j_m$, which means the score of $h^j_k$ is calculated by just weighting itself.

Subsequently, we use a softmax function to normalize $y^j\in R^{1\times W_iH_i}$ to obtain the normalized attention score matrix $A_i=[{a^i_1}^T,{a^i_2}^T,...,{a^i_{d}}^T]^T$, where $a^i\in R^{1\times W_iH_i}$. Considering the spatial pyramid, $a^i$ is calculated as follows,
\begin{gather}
  a^i = softmax(\mathscr{F}(y^{i-N+1},...,y^{i-1},y^{i})).
\label{equ:A_j}
\end{gather}
$\mathscr{F}$ is a fusion function for $y^j$ of all layers in the pyramid. Here we respectively investigate three fusion functions, element-wise maximum, element-wise sum and element-wise multiplication. Then, $a^i$ is L2 normalized \cite{ref91} to preserve ${a^i}^Ta^i=1$.

Finally, $A_i$ is used to aggregate the flattened feature maps $X_i$ of the $i$-th layer, to obtain a more discriminative representation $M_i$, as follows,
\begin{gather}
  M_i = A_i  X_i,~where~M_i\in R^{d\times C_i},~A_i\in R^{d\times W_iH_i},~X_i\in R^{W_iH_i\times C_i}.
\label{equ:M_i}
\end{gather}
\textbf{Discussion: } The attention mechanism extracts key features from the set of features by using attention scores to weight features while PCA extracts key features by using a set of basis vectors to project features. So, we give another insight on our self-attention process. $X_i$ can be also considered as stacked columns $X_i=[v^i_1,v^i_2,...,v^i_{C}]$ and $v^i\in R^{W_iH_i\times 1}$  represents a flattened global feature map divided by channel. We use PCA to generate the key features $M\in R^{d\times C_i}$ with principal dimensions from the stacked columns $\{v^i_m\}_{m=1}^{C}$. Let $S=[e_1, e_2,...,e_d]^T$ be the set of orthogonal basis vectors $\{e_n\}_{n=1}^d\in R^{W_iH_i\times 1}$, on which $v$ is projected for extracting principal components. So by PCA, we obtain $M_i=SX_i\in R^{d\times C_i}$, which is the same as the form of $M_i$ in Eq. \ref{equ:M_i}.

There exists a subtle correspondence between PCA and attention mechanism, namely the attention score $a^i\in R^{W_iH_i\times 1}$ corresponds to the basis vector $e\in R^{W_iH_i\times 1}$ even if the computing ways of $a$ and $e$ are different. In other words, this is an attention processing if we extract key features from $X_i$ considered as stacked channel-level features, as well as a PCA processing if extracting principal components from $X_i$ considered as stacked features divided by channels. Furthermore, ``$\oplus b_{attn_j}$" in Eq. \ref{equ:X_j} can be considered as the data central processing (subtracting mean) of PCA. Until now, our self-attention is actually a simplified version of PCA since it doesn't consider the interaction among features by just weighting itself to obtain attention scores. As described in the analysis of Eq. \ref{equ:X_j}, each attention score is obtained for $h$ by weighting $h$ itself. However in PCA, $S$ is usually obtained by eigenvalue decomposition of covariance matrix $X_iX_i^T$. Here we use another equivalent form \cite{ref89,ref90},
\begin{gather}
S=\mathop{\rm{argmin}}_S~-tr(SX_iX_i^TS^T),~where~S\in R^{d\times W_iH_i}, X_i\in R^{W_iH_i\times C_i},\notag\\
s.t.~SS^T=\textbf{I}.
\label{equ:pca}
\end{gather}
PCA uses the covariance matrix to obtain $S$, and in this way it utilizes the non-local interaction among features. Inspired by PCA, we add an interaction-aware loss item to generate an interaction-aware spatial pyramid attention layer, which use the non-local interaction information among channel features $h$ to further improve the effectiveness of self-attention. The details of interaction-aware loss are given in Section \ref{sec:sub:Design of Loss Function}.
\subsection{Temporal Aggregation}
\label{sec:sub:Temporal Aggregation}
Our interaction-aware spatial pyramid attention layer can accept not only a single image but also multiple frames as the input. We can easily extend it to a temporal architecture based on a original deep CNN. The obtained spatio-temporal pyramid attention network models the temporal sequences and  detects the key spatio-temporal information.

First, we sample $K$ frames from a video, and then input them into the network until the $i$-th layer to extract $K$ groups of feature maps $F_i\in R^{K \times W_{i} \times H_{i} \times C_{i}}$, which is flattened into $X_i\in R^{KW_iH_i\times C_i}$. To preserve the parameters and architecture of the network after the $i$-th layer unchanged, we seek to aggregate $F_i$ into a group of feature maps $M^{'}_i\in R^{W_{i} \times H_{i} \times C_{i}}$ which has the same size with $f_i$. To address this aggregation, we first construct feature pyramid $\{F_j\}|_{j=i-N+1}^{i}$ by $K$ frames. Then, we set $d=W_iH_i$, $\mathbb{W}_{attn_j}\in R^{d \times C_j}$ and $b_{attn_j}\in R^{d \times 1}$. It is interesting to note that $\mathbb{W}$ and $b$ of multiple frames have the same forms with that of a single frame in Eq. \ref{equ:X_j}. This is because we use $\mathbb{W}_{attn_j}$ to weight channels of feature maps. 
In addition, the size of output attention maps are fixed to $d=W_i\times H_i$, which is also the size of the $i$-th layer in the original CNN. So, as long as the feature maps of a CNN constructing the spatio-temporal pyramid are selected, the parameters are determined and are irrelevant to $K$. Based on this fact, we can expediently use frames of different numbers to train and test our networks. Finally, by replacing $f_j$ with $F_j$ and replacing $X_j\in R^{W_iH_i\times C_j}$ with $X_j\in R^{KW_iH_i\times C_j}$ in Eq. (\ref{equ:f_j})(\ref{equ:X_j})(\ref{equ:A_j})(\ref{equ:M_i}), we obtain the aggregated attention features $M$ and then it is reshaped to attention maps $M^{'}_i$.
\subsection{Design of Loss Function}
\label{sec:sub:Design of Loss Function}
We first present the specific form of our interaction-aware loss inspired by PCA. The original form of PCA is shown in Eq. \ref{equ:pca}. We use $A_i$ to represent $S$ and then  change this loss function into a derivable form as follows,
\begin{gather}
l_{interactive}=-\mathscr{X}((A_iX_iX_i^TA_i^T)\circ \textbf{I})+\mathscr{Y}((A_iA_i^T)\circ{(\textbf{1}-\textbf{I)}}),
\label{equ:pca_loss}
\end{gather}
where $\textbf{1}\in R^{d\times d}$ and $\textbf{I}\in R^{d\times d}$ are a matrix of all ones and an identity matrix respectively, $\circ$ is element-wise multiplication. $\mathscr{X}$ and $\mathscr{Y}$ are operations of sum of elements and quadratic sum of elements respectively. Minimizing the second term obtains the constraint $A_iA_i^T=\textbf{I}$ after the diagonal of $A_iA_i^T$ have already been all one, because $a_i$ is L2 normalized as descibed in Section \ref{sec:sub:Spatial Pyramid Attention Layer}.

In order to further enhance the representative ability of the attention maps, we propose a regularization about attention scores. Specifically, $a^i$ represents the attention scores calculated by all scales of the spatio-temporal pyramid and $softmax(y^j)$ represents the attention scores obtained by the $j$-th scale of the pyramid. To sufficiently utilize the information of different scales of the pyramid and make each scale focus on diverse parts as far as possible, we attempt to maximize the distance $\delta$ between $a^i$ and $softmax(y^j)$ or minimize the distance $1-\delta^2$. We define $\delta_{j,m,k}~(\in[0,1])$ as follows,
\begin{gather}
   \delta_{j,m,k} = ||a^i_{m,k}-softmax(y^j)_{m,k}||, m=1,...,d,k=1,...,KW_iH_i.
\label{equ:d_j}
\end{gather}
Subsequently, we give the specific form of our attention loss, as follows,
\begin{gather}
l_{attn}=~\sqrt{\sum_j\sum_m\sum_k(1-\delta_{j,m,k}^2)}.
\label{equ:attention_loss}
\end{gather}
We use the cross-entropy loss for the classification loss, and the final loss is formulated as follows,
\begin{gather}
   L = -\frac{1}{K}\sum_{t=1}^K\sum_{c=1}^{\mathfrak{C}}y_{t,c}log\hat{y}_{t,c}+\lambda\sum_{\theta}w_{\theta}^2+\beta l_{interactive}+\gamma l_{attn},
\label{equ:loss}
\end{gather}
where $y_t$ is one hot label vector, $\hat{y}_{t}$ is the vector of class probabilities obtained by the $t$-th frame. $\lambda$, $\beta$ and $\gamma$ are weight decay coefficients. $w_{\theta}$ are trainable weights of the network. $\mathfrak{C}$ is the class number.

\subsection{Implementation Details}
\begin{figure}[t]
\setlength{\abovecaptionskip}{-0mm} 
\setlength{\belowcaptionskip}{-0mm} 
\centering
\includegraphics[width=118mm,height=45mm]{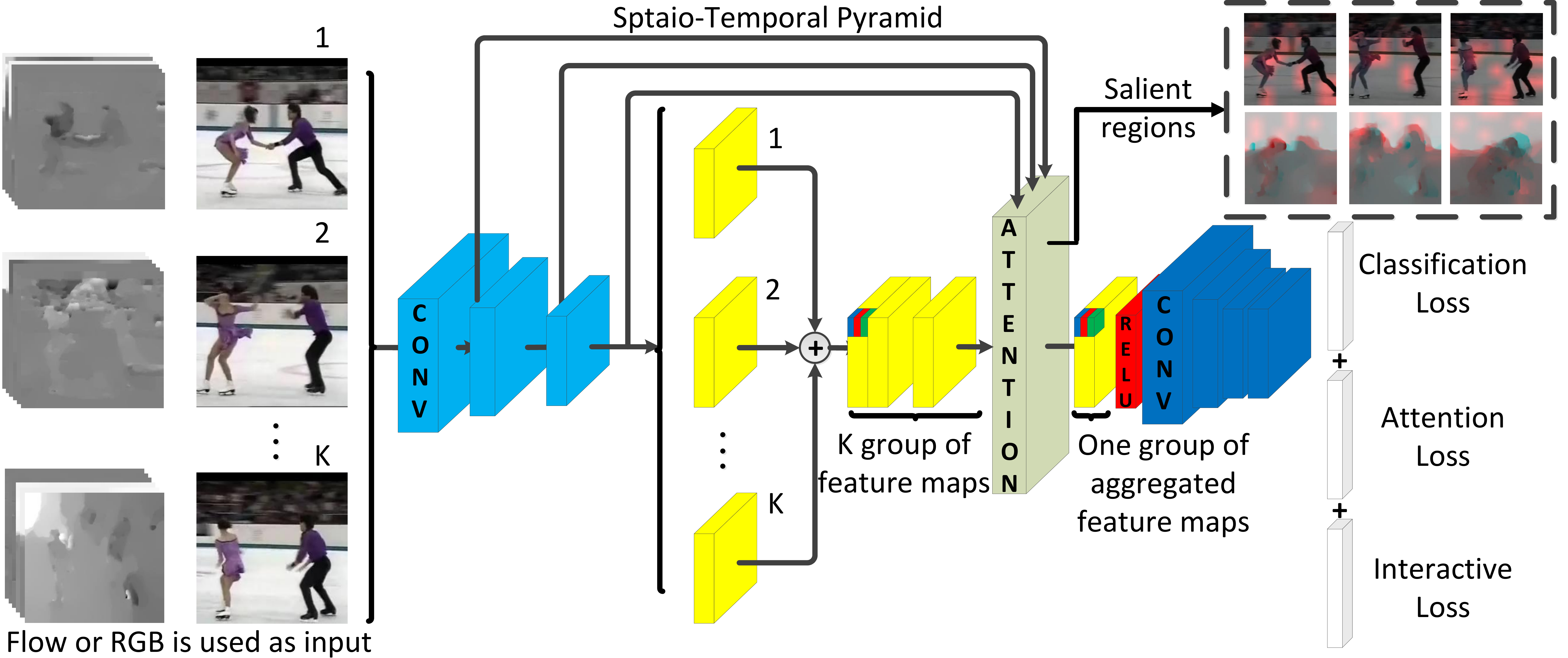}
\caption{The framework of the proposed network. An interaction-aware spatio-temporal pyramid attention layer is inserted to CNN to aggregate $K$ group of feature maps by $K$ frames. It utilizes multi-scale feature maps to accurately focus on the salient regions. }

\label{fig:whole}
\end{figure}
\textbf{Network architecture:} We incorporate interaction-aware spatio-temporal pyramid attention layer into general CNNs to form end-to-end attention networks for action classification. We employ Two-stream \cite{ref20} architecture and investigate VGGNet-16 \cite{ref45}, BN-Inception \cite{ref72} and Inception-ResNet-V2 \cite{ref46} respectively. The framework of the proposed attention network is shown in Figure \ref{fig:whole}. We use max pooling as the dawn sampling $\Re(\bullet)$ because max pooling is marginally superior than average pooling and convolution by our experiments. In addition, max pooling doesn't brings in any parameters. So we set max pooling as $\Re(\bullet)$.

\textbf{Training:} In RGB stream, we use the dropout of 0.5 with good performance. Since the spatial networks take RGB images as input, it is natural to use models trained on the ImageNet \cite{ref60} as initialization. In Flow stream, we use a dropout of 0.7 to avoid over-fitting on small flow datasets. Flow model is initialized by following \cite{ref64}. Data augmentation is done by performing random cropp/flipping of all the RGB and flow frames respectively. In practice, we follow the same segment setting \cite{ref48,ref68} for training, that is, we uniformly divide a video into 3 segments, choose a random frame from each segment and then form a $K=3$ sequences as the input of our networks. Furthermore, we also evaluate $K=1$ for training. We use mini-batch SGD to optimize our model with momentum of 0.9, weight decay $\lambda$ of 4$e^{-5}$, $\beta$ and $\gamma$ of 1$e^{-4}$, and a batch size of 64 for network training. We set the learning rate as follows. For spatial networks, the learning rate is initialized as 0.001 and decreases to its $1/10$ every 4000 iterations. The whole training procedure stops at 12000 iterations. For temporal networks, we initialize the learning rate as 0.005, which reduces to its $1/10$ after 12000 and 24000 iterations. The maximum iteration is set as 30000.

\textbf{Test:} During the test process on RGB and Flow streams, we investigate the effects of $K$ on the performance. We typically use $K=25$ frames for test and compare the results with other standard practice based methods. Furthermore, we investigate the performance of more frames ($>$25) in Section \ref{sec:seb:Evaluations of Temporal Aggregation}. We implement our models in TensorFlow \cite{ref71} with TITAN Xp$\times$2 GPUs.
\section{Experiments}
We evaluate our models on three challenging action classification benchmarks, the UCF101 \cite{ref55} , HMDB51 \cite{ref37} and untrimmed Charades \cite{ref62} datasets. For UCF101 and HMDB51, we follow the original evaluation scheme using three different training and test splits. We use the split 1 for ablation analysis and report the final performance by the average classification accuracy over these three splits. For Charades, we follow the evaluation pipeline of \cite{ref62}.
\subsection{Evaluations of the Proposed Attention Layer}
We investigate our interaction-aware spatio-temporal pyramid attention on the following five parts: (1) layer position of feature maps used for aggregation, (2) different fusion functions $\mathscr{F}$ of feature maps of pyramid, (3) numbers of layers in pyramid, (4) loss functions with ablated regularization items, and (5) the generality of our layer applied in different deep networks, including popular architectures VGGNet-16 \cite{ref45}, BN-Inception \cite{ref72} and Inception-ResNet-V2 \cite{ref46}.

In experiments about (1)-(4), if there is no special explanations, we choose Inception-ResNet-V2 as the baseline and typically choose the last layers of Inception-ResNet-A, B, and C without activation to construct a 3-scale interaction-aware spatio-temporal pyramid attention layer and optimize our networks without $l_{interactive}$ and $l_{attn}$. The selection of layers approximately satisfies the typical ratio of $2^n$ on the spatial size. The sizes of these three layers are $35 \times 35 \times 320$, $17 \times 17 \times 1088$ and $8 \times 8 \times 2080$. The experimental results are listed in Table \ref{tab:attn_layer}.

First, to determine which layer of feature maps are suitable to be aggregated with attention weights, we evaluate four different layers used for aggregation. Specifically, we respectively use the last layers of Inception-ResNet-A, B, and C without activation for aggregation. The pyramid is 1-scale and has only one layer for evaluation. In addition, we also evaluate the performance of using the last fully-connected layer and it is specially denoted as $X^{1536\times 1}(\mathbb{W}\in R^{1536\times 1},b\in R^{1536\times 1})$ for aggregation. The results are shown in Table \ref{tab:sub:which layer}, which clearly shows that the best performance is obtained by using the last convolutional layer of Block C. In general, the fully-connected layer loses much information of the spatial locations while feature maps of big spatial size is not very representative.

\begin{table}[t]
\footnotesize
\caption{Evaluations of (a) position of the top layer of pyramid; (b) different scales with Inception-ResNet-V2 (I-R-V2); (c) fusion functions with 3 scales;  (d) loss functions with I-R-V2 and our attention layer; and 3) our layer on VGGNet-16, BN-Inception and Inception-ResNet-V2. $K=3$ for training, K=25 for testing, on UCF101 split 1.}

\centering
\subtable[Position of the top layer of pyramid.]{
\label{tab:sub:which layer}
        \begin{tabular}{|c|c|c|}
        \hline
        Block (Inception-ResNet-V2) & RGB & Flow  \\
        \hline
        \hline
        Block A ($35\times35\times320$)  & 85.8\% &  83.5\% \\
        \hline
        Block B ($17 \times 17 \times 1088$) & 86.1\% &  83.7\% \\
        \hline
        Block C ($8 \times 8 \times 2080$) & \textbf{86.3\%} &  \textbf{84.0\%} \\
        \hline
        FC (1536) & 85.5\% &  83.4\% \\
        \hline
        \end{tabular}
}
\subtable[Different scales.]{
\label{tab:sub:scales}
        \begin{tabular}{|c|c|c|}
        \hline
        Scale & RGB & Flow  \\
        \hline
        \hline
        1 scale & 86.3\% &  84.0\% \\
        \hline
        2 scales & 86.8\% &  84.8\% \\
        \hline
        3 scales & \textbf{87.3\%} &  \textbf{85.5}\% \\
        \hline
        4 scales & 86.5\% &  85.0\% \\
        \hline
        \end{tabular}
}

\subtable[Performance of different fusion functions.]{
\label{tab:sub:function}
        \begin{tabular}{|c|c|}
        \hline
        Fusion Function~($\mathscr{F}$) & Accuracy \#RGB  \\
        \hline
        \hline
       Element-wise Maximum  & 85.7\%   \\
        \hline
       Element-wise Sum  & 86.4\%  \\
        \hline
       Element-wise  Multiplication  & \textbf{87.3\%} \\
        \hline
        \end{tabular}
}
\subtable[Loss functions with 3 scales.]{
\label{tab:sub:loss}
        \begin{tabular}{|c|c|c|c|}
        \hline
        Stream&$l_{inter}$&$l_{attn}$ & no~loss \\
        \hline
        \hline
        RGB&87.8\%&87.5\% & 87.3\% \\
        \hline
        Flow&86.1\%&85.7\%&85.5\% \\
        \hline
        Late fusion&94.7\%&94.4\%& 94.2\% \\
        \hline
        \end{tabular}
}
\subtable[Performance of the proposed attention layer on popular networks.]{
\label{tab:sub:nets}
        \begin{tabular}{|c|c|c|c|}
        \hline
        {Stream}/{($l_{inter}+l_{attn}$)} & VGGNet-16  & BN-Inception & Inception-ResNet-V2  \\
        \hline
        \hline
        RGB & 80.4\% & 84.5\%& 85.2\%\\
        RGB (3 scales) & \textbf{83.8\%} & \textbf{86.7\%} & \textbf{88.2\%} \\
        \hline
        Flow& 85.5\% & 87.2\%& 83.1\%  \\
        Flow (3 scales)& \textbf{87.1\%} & \textbf{87.9\%}& \textbf{86.5\%} \\
        \hline
        Late fusion& 90.7\% & 92.0\%& 92.6\%  \\
        Late fusion (3 scales)& \textbf{92.8\%} & \textbf{94.6\%} & \textbf{95.1\%}\\
        \hline
        \end{tabular}
}

\label{tab:attn_layer}
\end{table}

We explore different fusion functions $\mathscr{F}$ of feature maps of pyramid evaluated on RGB stream. Table \ref{tab:sub:function} lists the comparison results of different fusion strategies. Element-wise multiplication performs better than other candidate functions, and is therefore selected as a default fusion function. The similar conclusion is obtained in TLE \cite{ref68}, which uses element-wise multiplication to better encode feature maps of different frames.

Next, we respectively give the performance of our spatio-temporal pyramid attention layer of 1 scale (only using the top layer of pyramid), 2 scales (top 2 layers), 3 scales and 4 scales, as shown in Table \ref{tab:sub:scales}. The results using Inception-ResNet-V2 without our attention layer are shown in Table \ref{tab:sub:nets}. The results show that our 1-scale attention promotes 1.1\%/0.9\% performance on RGB/Flow and the performance can be further promoted by increasing the number of scales (3 scales promote 2.1\%/2.4\% on RGB/Flow). When we add the forth scale by using Conv2d\_4a\_3x3 ($71 \times 71 \times 92$ ) in Inception-ResNet-V2, the performance drops. It can be explained that spatio-temporal pyramid attention is indeed effective because more information from multi-scale feature maps of different receptive fields is considered to the aggregation of local features. However, when feature maps of bigger size are used, the receptive fields of these feature maps get narrow. Then, too local features of these narrow receptive fields will bring in the noise that leads to the drop of performance. We hence use the architecture of 3 layers (3 scales) to obtain the spatio-temporal pyramid in the following experiments.

Subsequently, we evaluate the effectiveness of our loss items $l_{interactive}$ and $l_{attn}$, as shown in  Table \ref{tab:sub:loss}. The results using two loss items are shown in Table \ref{tab:sub:nets}. The comparison results show that $l_{interactive}$ $+$ $l_{attn}$ are effective by improving the performance 0.9\%/1.0\%/0.9\% on RGB/Flow/Late fusion (3 scales). Furthermore, each of them also improves the performance individually compared to the results without these two loss items.

Finally, to investigate the generality of our attention layer, we respectively plug it into VGGNet-16, BN-Inception and Inception-Resnet-V2. The results are listed as shown in Table \ref{tab:sub:nets}. The late fusion approach means that the prediction scores of the RGB and Flow streams are averaged as the final video classification, as other methods \cite{ref20,ref48,ref62,ref68} do. For VGGNet-16, our attention network respectively promotes 3.4\%/1.6\%/2.1\% on RGB/Flow/Late fusion on the UCF101 split 1. For BN-Inception, our model respectively promotes 2.2\%/0.7\%/2.6\% on RGB/Flow/Late fusion. For Inception-Resnet-V2, our model respectively promotes 3.0\%/3.4\%/2.5\% on RGB/Flow/Late fusion. The improved results with our attention layer on these three deep networks prove the generality of our layer for general deep networks. The results without our layer are obtained by Two-Stream \cite{ref20} standard process.
\subsection{Evaluations of Temporal Aggregation}
\label{sec:seb:Evaluations of Temporal Aggregation}
\begin{figure}[!t]
  \centering
  \subfigure[$K$ of training]{
  \label{fig:frames:training}
    \includegraphics[width=57.4mm,height=38.5mm]{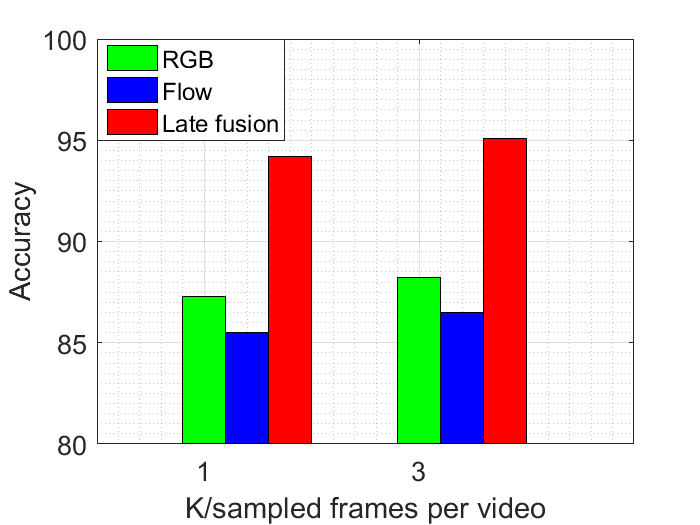}}
  \subfigure[$K$ for testing]{
  \label{fig:frames:testing}
    \includegraphics[width=57.4mm,height=38.5mm]{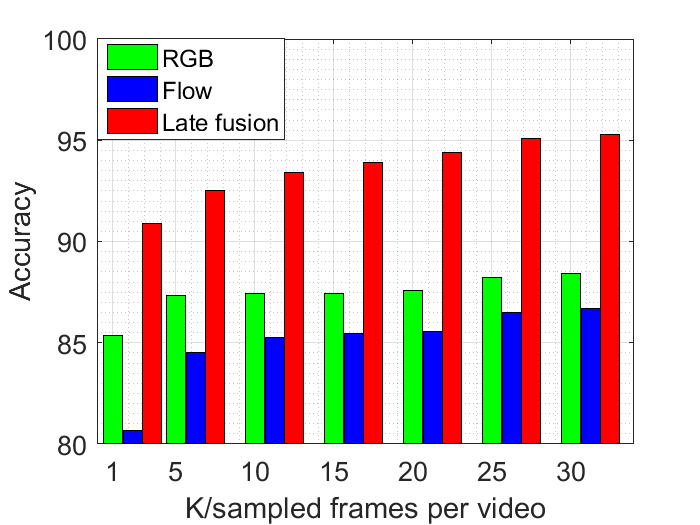}}
\caption{
Comparisons of different $K$ sampled frames per video for training and  testing process on UCF101 split1. We use 3-scale spatio-temporal pyramid interactive attention with Inception-ResNet-V2 to predict video.  Specifically, we respectively investigate $\{K=1,3\}$ frames for training and $\{K= 1,5,10,15,20,25,30\}$ frames for testing.}

\label{fig:frames}
\end{figure}
We investigate how the number $K$ of sampled frames for training and test affects our layer based on Inception-Resnet-V2. We first evaluate the performance with settings of $K=1,3$ for training and $K=25$ for test. When $K=1$, we randomly sample a frame from a video. When $K=3$, we randomly sample a frame from each of $K$ segments averagely divided from a video. The results are shown in Table \ref{fig:frames:training}. It can be seen training the model with temporal sequences contributes to the accuracy, as shown in Temporal Segment Networks \cite{ref48}. For fair comparison, we select $K=3$ for training in the later experiments.

Then, we evaluate $K$ for test, as shown in \ref{fig:frames:testing}. It can be seen that the performance gradually rises when more frames are sampled per video. Although more frames may bring irreverent information, our attention layer is able to select out the most effective information. We achieve the performance of 95.2\% when the most frames $K=30$ are selected because of the limited GPU memory. For further GPU of higher capacity, the performance of our model is probable to be better. For fair comparison, we typically show our results when $K=25$ frames \cite{ref20,ref48,ref64,ref62} for test are uniformly sampled from videos.

\begin{table}[!t]
\footnotesize
\setlength{\abovecaptionskip}{-0mm} 
\setlength{\belowcaptionskip}{-0mm} 
\begin{center}
\caption{
Comparisons with the state of the art on UCF101, HMDB51 and untrimmed Charades datasets. }
\subtable[Comparisons on  UCF101 and HMDB51  over 3 splits]{
\label{tab:sub:noidt}
\begin{tabular}{|c|c|c|}
\hline
    Algorithm & UCF101 & HMDB51  \\
    \hline
    \hline
    C3D \cite{ref61} & 85.2\% & -\ \\
    Soft Attention + LSTM \cite{ref74} & - & 41.3\% \\
    Two-Stream + LSTM \cite{ref63} & 88.6\% & -\ \\
    TDD+FV \cite{ref12} & 90.3\% & 63.2\% \\
    RNN+FV \cite{ref58} & 88.0\% & 54.3\% \\
    LTC \cite{ref42} & 91.7\% & 64.8\% \\
    ST-ResNet \cite{ref66} & 93.5\% & 66.4\% \\
    TSN (BN-Inception)  \cite{ref48} & 94.0\% & 68.5\% \\
    AdaScan \cite{ref67} & 89.4\% & 54.9\%\\
    ActionVLAD \cite{ref62} & 92.7\% & 66.9\% \\
    TLE (BN-Inception) \cite{ref68} & 95.6\% & 71.1\% \\
    Attention Cluster (ResNet-152) \cite{ref76} & 94.6\% & 69.2\% \\
    \hline
    \hline
    Ours (25 frames+BN-Inception)& 94.8\% & 69.6\% \\
    Ours (25 frames+Inception-ResNet-v2) & 95.3\% & 70.5\% \\
    \hline
    Ours (30 frames+Inception-ResNet-v2) & 95.5\% & 70.7\%\\
    \hline
    \end{tabular}
    }
 \subtable[Comparisons on the untrimmed Charades]{
\label{tab:sub:Charades}
\begin{tabular}{|c|c|c|}
\hline
        Algorithm&mAP&wAP \\
        \hline
        Two-stream + iDT (best reported) &18.6\%&-  \\
        RGB stream (BN-inception, TSN style training)&16.8\%&23.1\% \\
        ActionVLAD (RGB, BN-inception)&17.6\%&25.1\%\\   
        \hline
     & all losses/$l_{inter}$/$l_{attn}$/$no~loss$ & \\
        Ours (RGB, BN-Inception, 3~scales) & 20.2\%/19.8\%/18.7\%/18.3\%& 28.5\%\\
    \hline
    \end{tabular}
    }   
\end{center}
\end{table}
\subsection{Comparison to State of the Art}
In Table \ref{tab:sub:noidt}, we list recent state-of-the-art methods and comparable methods. We list our results based on two baselines of BN-Inception and Inception-Resnet-V2. It can be seen that the attention based method, such as Soft Attention + LSTM, is not very satisfying. Our method (BN-Inception) outperforms TSN (BN-Inception) by 0.8\%/1.1\% on UCF101/HMDB51 when the same $K=25$ uniformly sampled frames are used for evaluation. In addition, we obtain competitive performance comparing our method (BN-Inception, 94.8\%/69.6\%) with Attention Cluster (ResNet-152, 94.6\%/69.2\%) where ResNet-152 \cite{ref92,ref43} (single crop, 76.8\%) is more superior than BN-Inception \cite{ref72} (single crop, 74.8\%) with top-1 accuracy on ImageNet \cite{ref60}. We also add our model to Inception-ResNet-V2 and obtain the improved performance 95.3\%/70.5\%, which is comparable to the best performance with TLE. By sampling more frames per video, we further improve our performance to 95.5\%/70.7\%.  To prove the effectiveness, we further evaluate our model on the untrimmed Charades datasets in Table  \ref{tab:sub:Charades} by following the pipeline of another spatio-temporal aggregation method ActionVLAD [39]. Our model (3 scales+all losses) exceeds TSN 3.4\%(mAP)/5.4\%(weighted-AP/wAP) and ActionVLAD 2.6\%(mAP)/3.4\%(wAP). 

\subsection{Visualization Analysis}
We visualize what the proposed attention layer pays attention to over the frames and from different spatial positions. Specifically, let $l_{(k_j, w_j, h_j)}\in {R^{1\times C_i}}$ denote the vector in position ($k_j$, $w_j$, $h_j$) of $F_i\in R^{K \times W_{i} \times H_{i} \times C_{i}}$. Actually, $l_{(k_j, w_j, h_j)}$ is a local feature which describes a receptive field centered at the position $(w_j,h_j)$ in the $k_j$-th frame. By Equation \ref{equ:A_j}, $A_{w_m*h_m,k_j*w_j*h_j}\in R^{W_{i}H_{i} \times KW_{i}H_{i}}$ ($k_j\in[1,K]$,~$w_m,w_j\in[1,W_i]$,~$h_m,h_j\in[1,H_i]$) represents the attention score of $l_{(k_j, w_j, h_j)}$ contributing to spatial position ($w_m$, $h_m$) in the attention maps $M^{'}_i\in R^{W_{i} \times H_{i} \times C_{i}}$. We define the receptive fields with high attention scores as salient receptive fields and highlight them.
\begin{figure}[t]
  \centering
  \subfigure[Sampled RGB frames per video]{
    \label{} 
    \includegraphics[height=57mm,width=57mm]{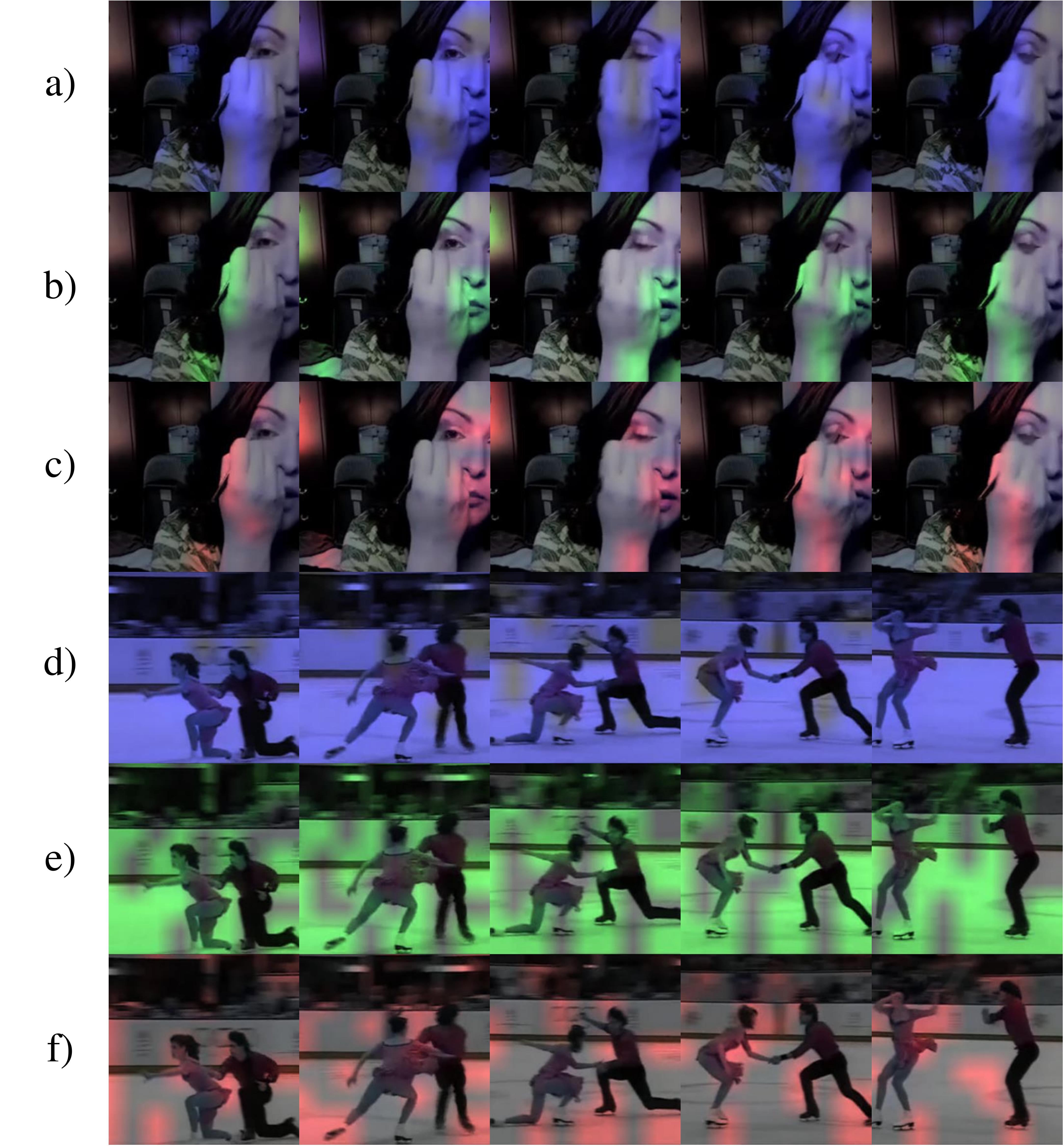}}
  \subfigure[Flow]{
    \label{} 
    \includegraphics[height=57mm,width=57mm]{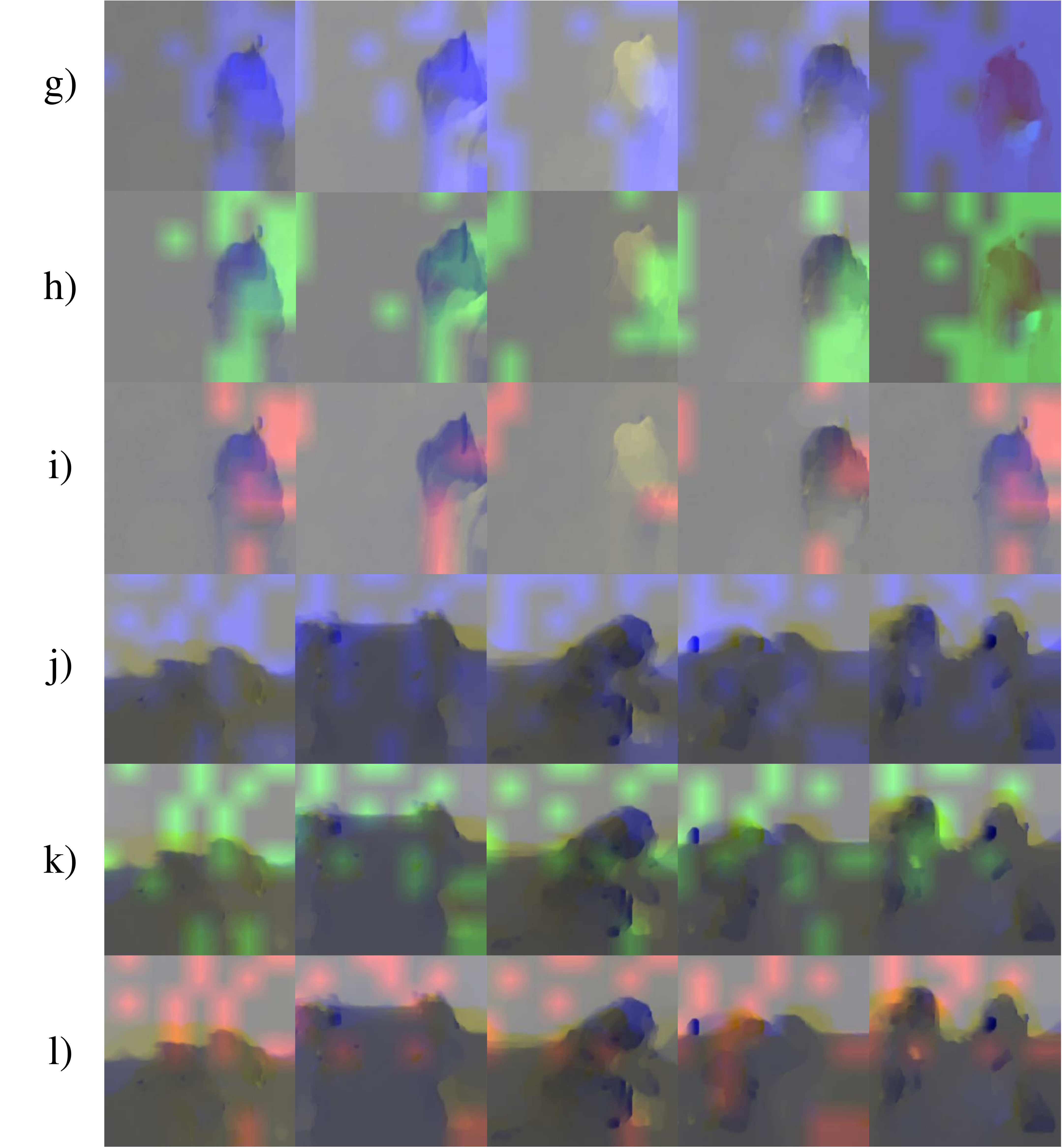}}
  \caption{Visualization of salient receptive fields in different frames from appearance (RGB) and motion (Flow) streams. Each row shows 5 frames from videos where blue,green and red regions respectively corresponds to the center of salient receptive fields by using 1 scale, 2 scales and 3 scales to obtain attention of different layers. Specifically, a)-c) and g)-i) respectively show the results of action 'ApplyEyeMakeup' from RGB and Flow. d)-f) and j)-l) show the results of action 'iceDancing'.}
\setlength{\abovecaptionskip}{-0mm} 
\setlength{\belowcaptionskip}{-0mm} 
\label{fig:sailent} 
\end{figure}
\begin{figure}[t]
\setlength{\abovecaptionskip}{-0mm} 
\setlength{\belowcaptionskip}{-0mm} 
\centering
\includegraphics[width=68mm,height=52mm]{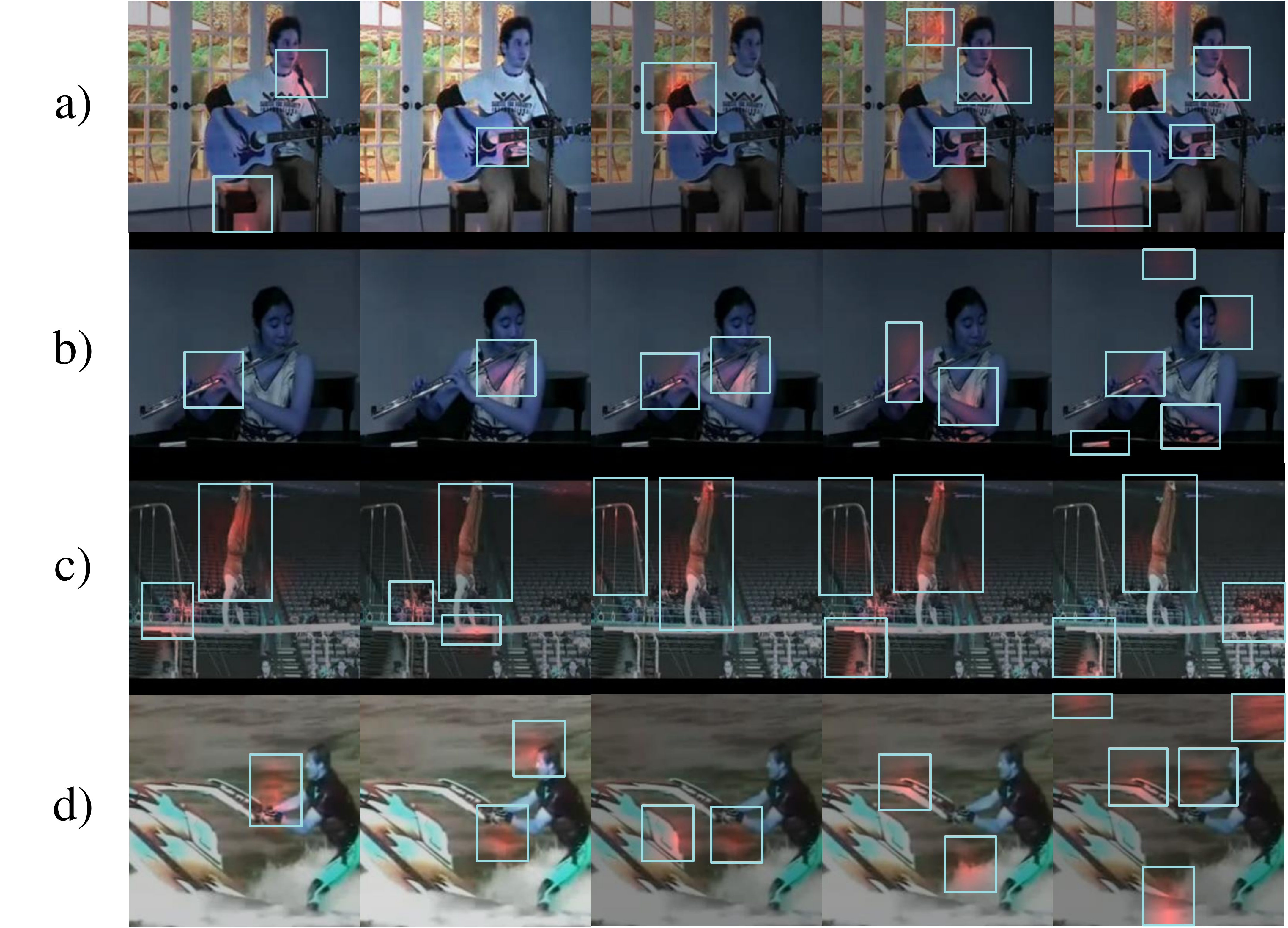}
\caption{
Visualization of salient receptive fields for different positions in attention feature maps from appearance (RGB) stream where 3-scale attention is used. Each column represents results of different positions ($w_m,h_m$). a) 'PlayingGuitar', b) 'PlayingFlute', c) 'ParallelBars', d) 'Skijet'.}
\label{fig:specific}
\end{figure}

Firstly, we visualize the salient receptive fields of $K$ input frames contributing to the fixed position $(w_m,h_m)$ in $F^{'}_i$. Namely for one frame, salient receptive fields centered at the positions satisfied $\{(w_j,h_j)|A_{w_m*h_m,k_j*w_j*h_j}>threshold,k_j=1,..,K,w_m=1, h_m=1\}$. Then we set a threshold 0.5 to show salient attention regions for 5 input frames, as shown in Figure \ref{fig:sailent}. The results show that our attention layer can pay attention to different key positions of actions over the frames. Moreover, we also shows the obtained salient regions by three methods which respectively use feature maps of 1 scale, 2 scales, and 3 scales to obtain the multi-scale attention scores $A$. It can be seen that the method using 3 scales pays attention to more specific and accurate action regions in every frame.

Secondly, for one fixed input frame we visualize the salient receptive fields contributing to different positions ($w_m,h_m$) in attention feature maps $F^{'}_i$. Namely, for one position, every salient spatial regions centered at the positions satisfied $\{(w_j,h_j)|A_{w_m*h_m,k_j*w_j*h_j}>threshold,k_j=1,w_m=1,..,W_i, h_m=1,..,H_i\}$. Then we set a higher threshold 0.7 for stronger discrimination, as shown in Figure \ref{fig:specific}. The results show that different positions ($w_m,h_m$) have different scopes of attention. From Figure \ref{fig:specific}{\color{red}(a)} it can be seen that action 'PlayingGuitar' is divided into multiple parts ('microphone', parts of 'guitar', hands) which are separately focused on from multiple positions in aggregated feature maps. The similar results can be seen in Figure \ref{fig:specific}{\color{red}(b)-(d)}. 

\section{Conclusion}
We have proposed an interaction-aware self-attention which is inspired by PCA to further use non-local information in feature maps. By constructing a spatial feature pyramid, our model improve the attention accuracy resulting in promoted classification accuracy. In addition, we have naturally extended our spatial model to a temporal model for action classification. The temporal model can accept changing input of any numbers and we have explored the influence of different training frames and test frames. We have investigated the performance of the proposed attention layer on three popular deep networks, VGG16, BN-Inception and Inception-ResNet-V2. The promoted performances have proven the generality of our model.

\section{Acknowledgments}
This work         is supported by the National Key R $\&$ D Plan (No. 2017YFB1002801, 2016QY01W0106), NSFC (Grant No. 61472420, 61472063, 61751212, 61472421, 61772225, U1736106, U1636218), the Key Research Program of Frontier Sciences, CAS (Grant No. XDB02070003, QYZDJ-SSW-JSC040), the CAS External cooperation key project, Research Project of State Grid General Aviation Company Limited (NO. 5201/2018-44001B), and Bing Li is also supported by Youth Innovation Promotion Association, CAS.
%
%
%
%
\clearpage

\bibliographystyle{splncs}
\bibliography{egbib}
\end{document}